\def\year{2022}\relax
\documentclass[letterpaper]{article} 
\usepackage{aaai22}  
\usepackage{times}  
\usepackage{helvet}  
\usepackage{courier}  
\usepackage[hyphens]{url}  
\usepackage{graphicx} 
\usepackage{natbib}  
\usepackage{caption} 
\DeclareCaptionStyle{ruled}{labelfont=normalfont,labelsep=colon,strut=off} 
\usepackage{algorithm}
\usepackage{algorithmic}

\usepackage{amsmath}
\usepackage[utf8]{inputenc} 
\usepackage{hyperref}       
\usepackage{url}            
\usepackage{booktabs}       
\usepackage{amsfonts}       
\usepackage{nicefrac}       
\usepackage{microtype}      
\usepackage{xcolor}   
\usepackage{wrapfig}
\usepackage{multirow}

\graphicspath{{./Fig/}}

  \DeclareSymbolFont{numbers}{T1}{ptm}{m}{n}
  \SetSymbolFont{numbers}{bold}{T1}{ptm}{bx}{n}
  \DeclareMathSymbol{0}\mathalpha{numbers}{"30}
  \DeclareMathSymbol{1}\mathalpha{numbers}{"31}
  \DeclareMathSymbol{2}\mathalpha{numbers}{"32}
  \DeclareMathSymbol{3}\mathalpha{numbers}{"33}
  \DeclareMathSymbol{4}\mathalpha{numbers}{"34}
  \DeclareMathSymbol{5}\mathalpha{numbers}{"35}
  \DeclareMathSymbol{6}\mathalpha{numbers}{"36}
  \DeclareMathSymbol{7}\mathalpha{numbers}{"37}
  \DeclareMathSymbol{8}\mathalpha{numbers}{"38}
  \DeclareMathSymbol{9}\mathalpha{numbers}{"39}

\usepackage{newfloat}
\usepackage{listings}
\floatstyle{ruled}
\newfloat{listing}{tb}{lst}{}
\floatname{listing}{Listing}
\title{Domain Knowledge-Based Automated Analog Circuit Design \\ with Deep Reinforcement Learning}
\author {
   Weidong Cao\textsuperscript{\rm 1,2},
   Mouhacine Benosman\textsuperscript{\rm 1},
    Xuan Zhang\textsuperscript{\rm 2},
    Rui Ma\textsuperscript{\rm 1}
}
\affiliations {
    \textsuperscript{\rm 1} Mitsubishi Electric Research Laboratories\\
    \textsuperscript{\rm 2} Department of Electrical \& Systems Engineering, Washington University in St. Louis\\
    \{weidong.cao, xuan.zhang\}@wustl.edu, \{benosman, rma\}@merl.com
}

\usepackage{bibentry}

\begin{document}

\maketitle

\begin{abstract}
The design automation of analog circuits is a longstanding challenge in the integrated circuit field. 
This paper presents a deep reinforcement learning method to expedite the design of analog circuits at the pre-layout stage, where the goal is to find device parameters to fulfill desired circuit specifications.
Our approach is inspired by experienced human designers who rely on domain knowledge of analog circuit design (e.g., circuit topology and couplings between circuit specifications) to tackle the problem.
Unlike all prior methods, our method originally incorporates such key domain knowledge into policy learning with a graph-based policy network, thereby best modeling the relations between circuit parameters and design targets.
Experimental results on exemplary circuits show it achieves human-level design accuracy ($\sim$\textbf{99}\%) with \textbf{1.5}$\times$ efficiency of existing best-performing methods.
Our method also shows better generalization ability to unseen specifications and optimality in circuit performance optimization.
Moreover, it applies to designing diverse analog circuits across different semiconductor technologies, breaking the limitations of prior ad-hoc methods in designing one particular type of analog circuits with conventional semiconductor technology.

\end{abstract}

\section{Introduction}
\label{sec:intro}

Recent advancements in deep learning have shown the great promise for transforming modern integrated circuit (IC) design workflows~\cite{bayesian,nvidia,google}.
By formulating each design stage as a learning problem, the IC development cycles can be significantly shortened compared to the manners with conventional Electronic Design Automation (EDA) tools.
For example, Google~\cite{google} and Nvidia~\cite{nvidia} have shown that deep learning methods can improve the design efficiency at the order of $\sim$100$\times$ in some stages of digital ICs, such as floorplanning and power estimation.
Analog circuits are a critical type of ICs to connect our physical analog world and modern digital information world~\cite{fangxu}.
Unlike digital ICs following standard design flows with EDA tools or emerging highly-efficient learning-based design automation methods, analog circuit design relies on onerous human efforts and lacks effective design automation techniques at all stages~\cite{bayesian,GNN_distributed, Neu_ADC_DATE,Pipeline_ICCAD}.

Pre-layout design is one key stage in analog circuit design flow. 
It can be represented as a parameter-to-specification (P2S) optimization problem.
The goal is to find optimal device parameters (e.g., width and finger number of transistors) to meet desired circuit specifications (e.g., power and bandwidth) based on a pre-determined circuit topology.
This problem is very challenging as it seeks optimum parameters of diverse devices in a huge design space without exact rules.
Even worse, the searching complexity grows exponentially with the increase of both design parameters and desired circuit specifications.
Conventionally, human designers leverage rich domain knowledge, e.g., topologies of circuits and couplings of specifications~\cite{bayesian,GNN_distributed}, to manually derive device parameters.
Particularly, a human designer pays intense efforts to obtain empirical equations between device parameters and circuit specifications based on a simplified circuit topology.
However, due to model simplicity and couplings between circuit specifications, tens and even hundreds of iterative fine tunings are required to ensure the design accuracy and reliability.

In this paper, we propose a reinforcement learning (RL) method for the P2S optimization problem, where an intelligent RL agent can autonomously figure out optimal device parameters for the desired circuit specifications.
Unlike all prior arts~\cite{GNN_distributed,op_amp_FCNN,autockt_berke,RL_1,GCN_RL_MIT}, our approach originally incorporates the key domain knowledge of analog circuit design into the learning framework, thereby reaching human-level design accuracy ($\sim$\textbf{99}\%) with \textbf{1.5}$\times$ efficiency of existing best-performing methods~\cite{autockt_berke,RL_1,GCN_RL_MIT}.
Even for a few failed cases, the RL agent can still provide hints to warm start a manual tuning method to ensure 100\% design accuracy.
Its great ability is enabled by the tailored policy network composed of a graph neural network (GNN) and a fully connected neural network (FCNN).
The GNN is built upon the full topology of a given circuit.
It can capture the underlying physics of the circuit, e.g., device's parameters, connections, and interactions.
The FCNN extracts the couplings of circuit specifications.
With such a unique policy network, our RL agent learns the best policy by incorporating key domain knowledge into training and makes optimal sequential decisions like a human expert to find device parameters.

Our method also breaks the limitations of prior ad-hoc methods~\cite{autockt_berke,RL_1,GCN_RL_MIT} in designing one type of low-frequency analog circuits with conventional complementary metal-oxide semiconductor (CMOS) technology.
It leverages transfer leaning and exploits commonly essential physical features of circuits, thereby applying to design various analog circuits across different semiconductor technologies.
Particularly, it can well design radio-frequency (RF) power electronic circuits (PECs) with emerging gallium nitride (GaN) technology~\cite{Gallium_nitride}.
RF PECs are a subclass of analog circuits that specially deal with high power densities and high-frequency signals, demanding more sophisticated analyses besides the design challenges faced by low-frequency analog circuits~\cite{bayesian}. 
Our work shows that RL methods combined with domain knowledge of circuit design are promising to bring us closer to a future where IC designers can be assisted by artificial agents with massive circuitry optimization experiences.

\section{Related Work}
\label{sec:re_work}
\noindent{\textbf{Existing Design Automation Methodologies:}}
Various design automation techniques have been proposed for the P2S problem of analog circuits, mainly falling into two categories: optimization-based methods and learning-based methods.
Optimization-based methods include Bayesian Optimization~\cite{bayesian}, Geometric Programming~\cite{geo_programming}, Genetic Algorithms~\cite{genetic}, and Simulated Annealing~\cite{simulated}.
These methods use corresponding algorithms to search for optimal device parameters.
However, they suffer from several key issues, such as divergence, being stuck at a local optimum, low-effectiveness for circuits with relatively large design space, and requiring to re-start from scratch if any change is made on the given specifications.

Learning-based methods have recently emerged to overcome the limitations of the optimization-based methods. 
Supervised learning methods~\cite{GNN_distributed,op_amp_FCNN} learn the static mapping between device parameters and circuit specifications.
Due to the inherent approximation errors, they cannot ensure high design accuracy and endure weak generalization abilities.
RL methods~\cite{autockt_berke,RL_1,GCN_RL_MIT} learn an optimal policy from the state space of a circuit to the action space of device parameters, which are solving a dynamic programming problem.
They often achieve higher design accuracy and stronger generalization abilities.

Despite such promises, existing RL methods~\cite{autockt_berke,RL_1,GCN_RL_MIT} are unable to reach human-level design accuracy, i.e., $\sim$100\%.
It could be attributed to the fact that none of them capture the key domain knowledge of analog IC design (e.g., topologies of circuits and couplings of specifications) into policy training, thereby failing to accurately learn the complex relations between device parameters and circuit specifications and leading to sub-optimal policies.
There is a prior RL method~\cite{GCN_RL_MIT} using a graph convolutional network (GCN) to process a circuit topology graph but with two key issues.
First, only a partial circuit topology is adopted by excluding power supply and bias nodes which however are the indispensable parts of a circuit graph. 
Second, the node features in the GCN are all static technology information, such as threshold voltage and electron mobility.
Without including the essential dynamic (variable) device parameters into node features, it is hard to learn the subtle relations between device parameters and circuit specifications.
Moreover, these prior arts are limited to design only low-frequency CMOS analog circuits, i.e., operational amplifiers (Op-Amp).
They are not readily for designing more advanced analog circuits, e.g., RF circuits, which require much more time-consuming characterizations.
Without overcoming the issue, a much longer training time is needed by them before used for inference/deployment.
Our method is inspired by human experts, which takes in key domain knowledge and the most essential features of analog circuits across different semiconductor technologies (e.g., CMOS and GaN).
With such key observations as well as the superior optimization ability and transfer learning ability of RL, it applies to design diverse analog circuits (including RF circuits) with human-level accuracy and higher efficiency.

\begin{figure*}[!t]
\begin{center}
\centerline{\includegraphics[width=0.75\linewidth]{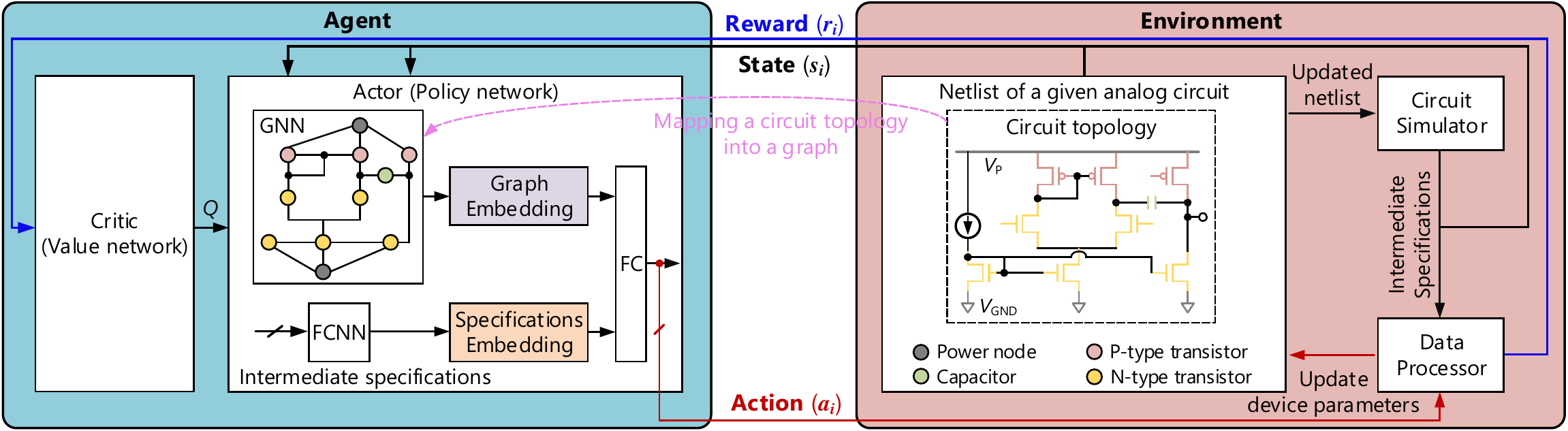}}
\caption{Overview of our RL framework for automated design of analog circuits. The RL agent is based on an actor-critic method. The environment consists of a netlist of any analog circuit with a given topology, a circuit simulator, and a data processor. At each time step $i$, the agent outputs an action $a_i$ to update device parameters with its policy network according to the state $s_i$ and then receives the reward $r_i$ from the environment. Our policy network is composed of a circuit-topology-based GNN and an FCNN. Here, we use a two-stage Op-Amp as an example to show how to map a circuit topology into a graph.}
\label{fig: overview}
\end{center}
\vskip -0.2in
\end{figure*}

\noindent{\textbf{Learning with Graph Neural Networks:}}
Graph neural networks, e.g., GCN~\cite{gcn} and graph attention network (GAT)~\cite{gan}, are emerging neural networks directly operating on non-Euclidean data structure resembling graphs.
They have gained increasing popularity in various domains, including social network~\cite{gnn_social} and recommendation system~\cite{graph_recommendation}.
Researchers have recently applied GNN to model circuit structure.
For example, a prior work~\cite{GNN_distributed} shows that the electromagnetic properties of distributed circuits can be learned in a supervised learning manner with a GNN.
Our work harnesses GCN and GAT to capture the physics of a given circuit, e.g., device's parameters
and interactions, for our policy network.
We show that a GAT with multi-head attention mechanism on nodes can better learn high-dimensional complex relations between circuit nodes.
\section{Approach}
\label{sec:RL}


We propose an RL approach for the P2S problem of analog circuit design at the pre-layout stage. It overview is shown in Figure~\ref{fig: overview}.
In out setting, the RL agent starts from an initial state $s_0$ with a group of initial device parameters and a group of randomly sampled desired specifications.
At each time step $i$, the agent observes state $s_i$ and takes action ${a}_i$ to update all device parameters for the given circuit based on the policy.
It then arrives at a new state $s_{i+1}$ and receives a reward $r_i$ from the environment.
The termination of an episode is that the design goals are realized or a pre-defined maximum step is reached. 
The agent iterates through the episode with multiple steps and accumulates the reward at each step to obtain the final return.
With multiple such episodes, the agent improves its decision quality and finally learns the best policy to maximize the return.
Next, we define the reward $r$, action $a$, state $s$, environment, the policy network $\pi_{\theta}(a|s)$ parameterized by $\theta$, and the optimization method used to train these parameters, and finally the transfer learning method to improve the training speed and design efficiency of RF power circuits.

\noindent{\textbf{Reward Function:}} The reward is directly related to the design goal.
We define the reward $r_i$ at each time step $i$ as 
\begin{equation}
 r_i= r, ~~\text{if} ~r < 0 ~~~ \text{or} ~~~  r_i=R, ~~\text{if} ~r =0,
\label{eq:reward}
\end{equation}
where $r=\sum\nolimits_{j=0}^{N-1}\min\{{(g^j_i-g^j_{*})}/{(g^j_i+g^j_{*})},0\}$ is a normalized difference between the intermediate specifications $g_i$ and the given specifications $g_{*}$.
The upper bound of $r$ is set to be 0 to avoid over-optimizing the parameters once the given specifications are reached.
All $N$ specifications are equally important.
We also give a large reward (i.e., $R=10$) to encourage the agent if the design goals are reached at some step.
The episode return $R_{s_0,g_{*}}$ of searching optimal device parameters for the given goals $g^{}_{*}$ starting from an initial state $s_0$, is the accumulated reward of all steps: $R_{s_0,g_{*}}= \sum\nolimits_{i=0} r_i$.

\noindent{\textbf{Action Representation:}} Inspired by human designers who iterate with fine-grained tuning steps to find optimal device parameters, we use discrete action space to  tune device parameters.
For each tunable parameter $x$ of a device (e.g., width and finger number of transistors), there are three possible actions at each step: increasing ($x+\triangle x$), keeping ($x+0$), or decreasing ($x-\triangle x$) the parameter, where ``$\triangle x$" is the smallest unit to update the parameter within its bound $[x_{\min},x_{\max}]$.
Assuming total $M$ device parameters, the output of the policy network is an $M\times 3$ probability distribution matrix with each row corresponding to a parameter.
The action is taken based on the probability distribution.

\noindent{\textbf{Environment:}}
A circuit design environment is used in this work.
It consists of a given circuit netlist, an industrial circuit simulator, such as Cadence Spectre for low-frequency circuits or Keysight Advanced Design system (ADS) for high-frequency power electronic circuits, and a data processor.
As shown in Figure~\ref{fig: overview}, the simulator obtains intermediate circuit specifications at each time step.
The data processor then deals with the simulated results to feed back a reward to the agent using Eq.~\eqref{eq:reward}.
Meanwhile, it updates the device parameters to rewrite the circuit netlist based on the actions from the agent.

\noindent{\textbf{State Representation:}}
Capturing critical and adequate domain knowledge from the environment is key to training a good RL agent. 
In a circuit design environment, the circuit itself and the intermediate specifications are the main domain observations.
In our work, we for the first time adopt these two key practical observations to represent each state $s_i$.
We use a graph $G(V,E)$ to model the circuit based on its topology, where each node in set $V$ is a device and the connections between devices form the edge set $E$.
We also treat the power supply ($V_{\text{P}}$), ground ($V_{\text{GND}}$), and other DC bias voltages as extra nodes.
To show how to map a circuit topology into a graph, Figure~\ref{fig: overview} takes a two-stage Op-Amp as an example.
For a circuit with $n$ nodes, the state for the $k^{\text{th}}$ node is defined as its node feature $(t,\vec{p})$, where $t$ is the binary representation of the node type and $\vec{p}$ is the parameter vector of the node.
For transistors, the parameters are the width ($x_{\text{W}}$) and the finger number ($x_{\text{F}}$) while for capacitors, resistors, and inductors, the parameter is the scalar value of each device.
The parameter for power supply (ground or DC bias) is a voltage of $V_{\text{P}}$ (0 for $V_{\text{GND}}$ or $V_{\text{bias,}k}$ for bias node $k$).
Zero padding is used to ensure that the length of $\vec{p}$ for each node is the same. 
For a circuit with five different types of devices, two power nodes, one bias, the state of an N-type transistor is $[{0,0,1}, {x_{\text{W}},x_{\text{F}}}]$.
We also create a vector to represent intermediate specifications. 
For example, to design the Op-Amp, the state vector of specifications is expressed as $[G, B,PM,P]$ which are 
\text{gain} ($G$), \text{bandwidth} ($B$), \text{phase margin} ($PM$), and \text{power consumption} ($P$).

\noindent{\textbf{Agent:}} Experienced human designers mainly rely on the domain knowledge, i.e., topologies of circuits and couplings of circuit specifications to tune device parameters as these factors dominate the relations between device parameters and circuit specifications.
A good RL agent is expected to be able to incorporate the key domain knowledge into policy learning such that it can make human-level decisions.
To enable such an ability of our agent (based on actor-critic method~\cite{actor_critic}), we propose a novel policy network as shown in Figure~\ref{fig: overview}.
It consists of a circuit-topology-based GNN and an FCNN, which is termed GNN-FC-based policy network.
The role of the GNN is to distill the underlying physics (e.g., device’s types, parameters, and interactions) of a circuit graph into low-dimensional vector embedding,
which clearly differs from the one of prior work~\cite{GCN_RL_MIT}] in structure, state encoding, and physical insights.
While the FCNN takes the design goals as inputs to extract their coupled relations, i.e., design trade-offs.
The graph embedding and the FCNN embedding are then concatenated for further processing by the final fully-connected (FC) layers to update the actions.
Combining all these parts forms the policy network $\pi_{\theta}(a|s)$ parameterized by $\theta$.
The value network preserves the same architecture as the policy network except of the last layer.
It evaluates the actor's decision quality by yielding an estimation of the expected reward, $Q$, for the current policy execution.
Our goal is to make the RL agent gain rich circuit design experiences and generate high-quality decisions by interacting with the environment.
The objective function of the problem can be formally defined as $J(\theta, G)={1}/{H} \cdot \sum\nolimits_{g\sim G}\mathbb{E}_{g,s\sim \pi_{\theta}}[R_{s,g}]$.
Here, $H$ is the the space size of all desired specifications $G$ and $R_{s,g}$ is the episode reward.
Given the cumulative reward for each episode, we use Proximal Policy Optimization (PPO)~\cite{ppo} to train the policy network.

\begin{table*}[!t]
\caption{Design space of device parameters and sampling space of desired specifications of two circuit benchmarks.}
\begin{center}
\begin{scriptsize}
\vskip -0.15in
\begin{tabular}{c|cccc|cc}
\toprule
Circuit types                                                       & \multicolumn{4}{c|}{Two-stage Op-Amp} & \multicolumn{2}{c}{RF PA} \\ 
\hline
Implementation technology & \multicolumn{4}{c|}{45 nm CMOS}       & \multicolumn{2}{c}{150 nm GaN}           \\ \hline
\# of device parameters  & \multicolumn{4}{c|}{$2\cdot7 + 1 =15$}      & \multicolumn{2}{c}{$2\cdot7=14$}               \\ \hline
\begin{tabular}[c]{@{}c@{}}Parameter constraints\\ (Design space) \end{tabular} &
  \begin{tabular}[c]{@{}c@{}}Width ($\mu$m)\\ $[1,100]$\end{tabular} &
  \begin{tabular}[c]{@{}c@{}}\# of fingers\\ $[2, 32]$\end{tabular} &
  \multicolumn{2}{c|}{\begin{tabular}[c]{@{}c@{}}capacitance (\textit{p}F)\\ $[0.1, 10]$\end{tabular}} &
  \begin{tabular}[c]{@{}c@{}}Width ($\mu$m)\\ $[16,100]$\end{tabular} &
  \begin{tabular}[c]{@{}c@{}}\# of fingers\\ $1, 2,...,16$\end{tabular} \\ \hline
\begin{tabular}[c]{@{}c@{}}Desired specifications\\ (Sampling space)\end{tabular} &
  \begin{tabular}[c]{@{}c@{}}Gain ($G$)\\ $[300, 500]$\end{tabular} &
  \begin{tabular}[c]{@{}c@{}}Bandwidth ($B$)\\ $[10^6, 2.5\cdot 10^7]$ Hz\end{tabular} &
  \begin{tabular}[c]{@{}c@{}}Phase margin ($PM$)\\  $[55^{\circ}, 60^{\circ}]$\end{tabular} &
  \begin{tabular}[c]{@{}c@{}}Power consumption ($P$)\\ $[10^{-4}, 10^{-2}]$ W\end{tabular} &
  \begin{tabular}[c]{@{}c@{}}Power efficiency ($E$)\\ $[50\%, 60\%]$\end{tabular} &
  \begin{tabular}[c]{@{}c@{}}Output power ($P$) \\ $[2,3]$ W\end{tabular} \\   \bottomrule
\end{tabular}
\end{scriptsize}
\end{center}
\label{tab1}
\vskip -0.10in
\end{table*}

\begin{figure*}[!t]
\begin{center}
\centerline{\includegraphics[width=0.75 \linewidth]{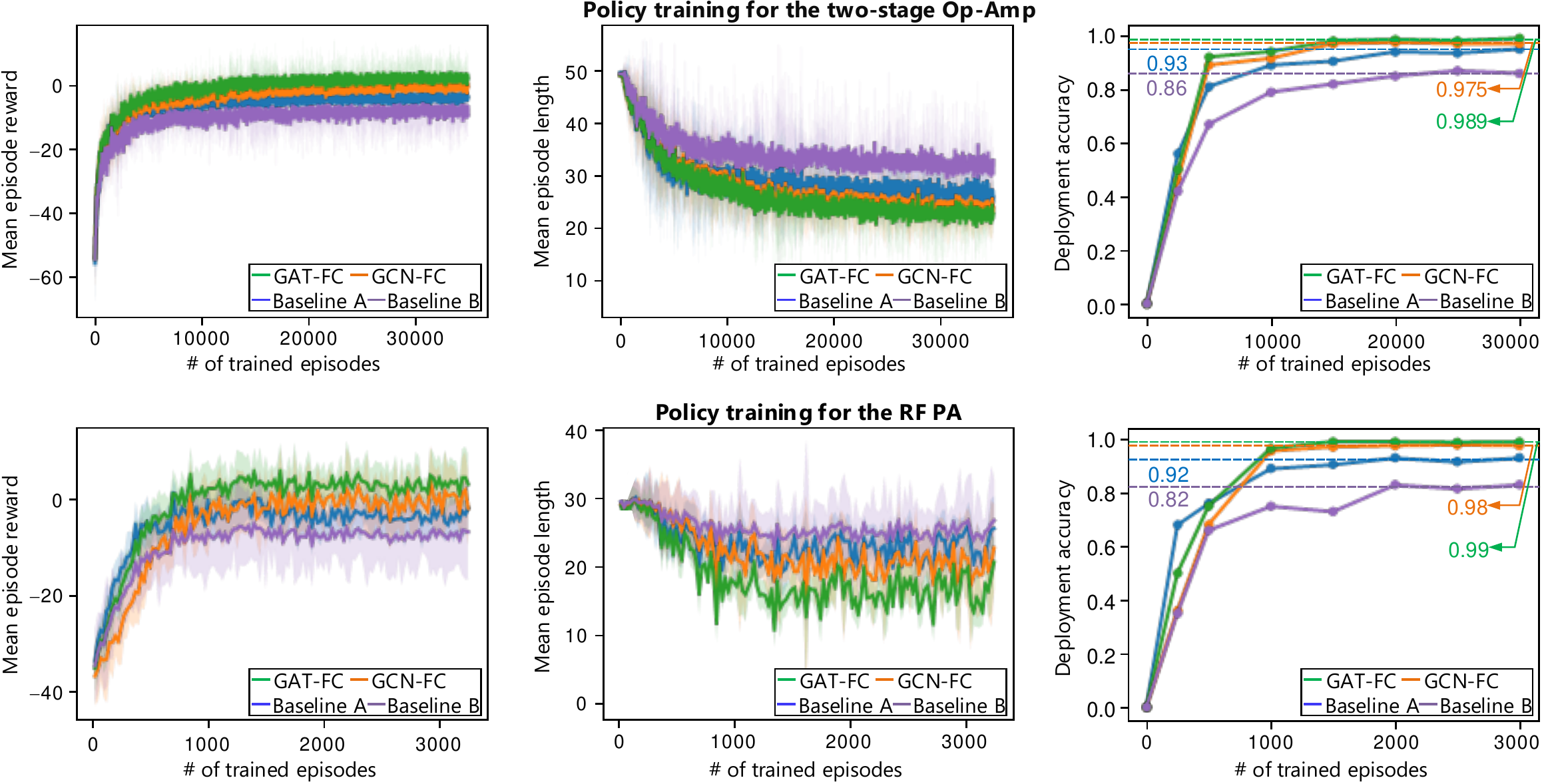}}
\caption{Evolution of mean episode reward, mean episode length, and deployment accuracy of RL agents w.r.t. the training episodes. Two rows correspond to the two-stage Op-Amp and the RF PA, respectively. All results are based on 6 random seeds.}
\label{fig: training_comp}
\end{center}
\vskip -0.25in
\end{figure*}

\noindent{\textbf{Transfer Learning:}}
We use transfer learning to speed up RF circuit design.
Generally, AC and DC simulations are enough to obtain all intermediate specifications $g_i$ at time step $i$ for low-frequency analog circuits.
Such simulations are fast within tens of milliseconds by using Cadence Spectre, which does not delay the learning of RL agents.
However, RF power circuits (e.g., RF power amplifiers) often require more sophisticated  simulations to obtain accurate intermediate specifications which is timing-consuming.
Typically, one needs to use Harmonic Balance (HB) simulation ($\sim$1 minute/round in ADS) to attain intermediate specifications.
It significantly delays the reward calculation and the training of RL agents.
To tackle this issue, fast ($\sim$1 second) but rough DC simulation is used to replace HB simulation.
It can obtain the not-very-accurate intermediate specifications for the quick approximation of the reward.
Our analyses show that the approximated rewards are often in $\pm$10\% error range compared to the ones obtained from the HB simulation.
Therefore, the learning process is remarkably speeded up. 
However, during the deployment (inference) stage for design automation, we still use HB simulation to guarantee the design quality and reliability.
In this way, the learned experiences from a coarse simulation environment can be well transferred into a fine simulation environment as verified by our results.
We think this may be due to the fact that a coarse design environment also provides sufficient information for the RL agent to learn the complicated relation between the device parameters and specifications.
For other advanced analog circuits, similar approximated rewards can also be obtained correspondingly.


\section{Experiments}
\label{sec:exp}

Two representative analog circuits are used to evaluate all methods.
First, the CMOS two-stage Op-Amp shown in Figure~\ref{fig: overview} is used as a low-frequency example.
It is a standard benchmark taken by all prior methods~\cite{bayesian, autockt_berke,RL_1, GCN_RL_MIT,genetic}. 
Second, a GaN RF power amplifier (PA)~\cite{polar_trans} whose schematic is not shown here is used as a more challenging high-frequency example.
The design space of device parameters and the sampling space of desired specifications for the two circuits are listed in Table~\ref{tab1}.
Due to the hard constraints imposed by the practical circuit design, there are total 15 and 14 parameters for the Op-Amp and RF PA.

Prior RL methods~\cite{autockt_berke,RL_1,GCN_RL_MIT} are mainly used to compare with ours.
These prior RL methods exclude the key domain knowledge into policy learning and are not capable of designing RF circuits.
Despite of different technical details in many aspects, they can be classified into two baselines.
Baseline A includes the prior work~\cite{autockt_berke,RL_1} which focuses on P2S optimization.
It simply observes intermediate device parameters, intermediate and given specifications from the environment and vectorizes them to train a feedforward policy work.
Baseline B is the prior work~\cite{GCN_RL_MIT} which aims to optimize the figure-of-merit (FoM), i.e., finding device parameters to attain the best overall performance for an analog circuit.
As introduced before, it uses all static semiconductor technology information as observations to train a partial-circuit-topology-based policy network.
Such a method is often found to be divergent during training.

For conservative comparisons, we interpret and implement these RL arts~\cite{autockt_berke,RL_1,GCN_RL_MIT} with our method.
First, we use the GCN design in our policy network as a more advanced implementation for the Baseline B.
Note that our GCN part is not only built upon a full circuit topology but also uses the essential dynamic (variable) device parameters as node features to better learn the relations between device parameters and circuit specifications.
Second, we build these RL baselines with the PPO technique~\cite{ppo} and discrete action space as done in our work as well as the transfer leaning technique to enable them to design RF circuits.
In our methods, GCN~\cite{gcn} and GAT~\cite{gan} are used as the GNN part to capture the underlying physics of a full circuit topology.
Therefore, our methods have two versions: GCN-FC policy and GAT-FC policy.
Compared to the GCN, the GAT has a multi-head attention mechanism on graph nodes, which can help to extract higher-dimensional interactions between circuitry nodes.
We compare these methods in the context of two applications, i.e., P2S problem~\cite{autockt_berke,RL_1} and FoM optimization~\cite{GCN_RL_MIT}.
All our experiments are performed on an 8-core Intel CPU.
We train separate RL agent for each circuit.
We use equal amount of network parameters and the same set-ups for each baseline.
Additionally, surprised learning method~\cite{op_amp_FCNN}, Genetic Algorithm~\cite{genetic}, and Bayesian Optimization~\cite{bayesian} are also used as auxiliary comparisons with our method.
We also use GAT to implement Baseline B. 
All these results are summarized in Table~\ref{tab2}.


\noindent{\textbf{P2S Optimization:}}
Figure~\ref{fig: training_comp} shows the training curves (i.e., mean episode reward, mean episode length, and deployment accuracy) of RL agents implemented with different RL methods for the P2S optimization problem.
The maximum episode length for each Op-Amp agent (RF PA agent) is set to be 50 (30). 
The total episodes used to train the two RL agents are chosen to be $3.5\cdot10^4$ and $3.5\cdot10^3$, respectively. 
As observed, our method achieves higher reward (left column), shorter mean episode length (middle column), and higher deployment accuracy (right column) than all RL baselines.
Policy deployment applies a trained policy to automatically find the device parameters for given specifications.
Each point in Figure~\ref{fig: training_comp} (right column) comes from deploying each RL agent for 200 groups of randomly sampled specifications in Table~\ref{tab1}.
The comparison shows that the our methods achieve higher design efficiency (with fewer deployment steps per episode) and human-level design accuracy (i.e., $99\%$ policy deployment accuracy) for both circuits design.
Particularly, we also note that the GAT-FC-based policy is superior to the GCN-FC-based policy.
Such a comparison shows that circuit topology is an important ingredient in RL-based policy learning.
And a better circuit topology modeling method, that is using GAT with multi-head attention mechanism to learn higher-dimensional interactions among circuitry nodes, can further improve the performance of a policy.


\begin{figure*}[!t]
\begin{center}
\centerline{\includegraphics[width=0.75 \linewidth]{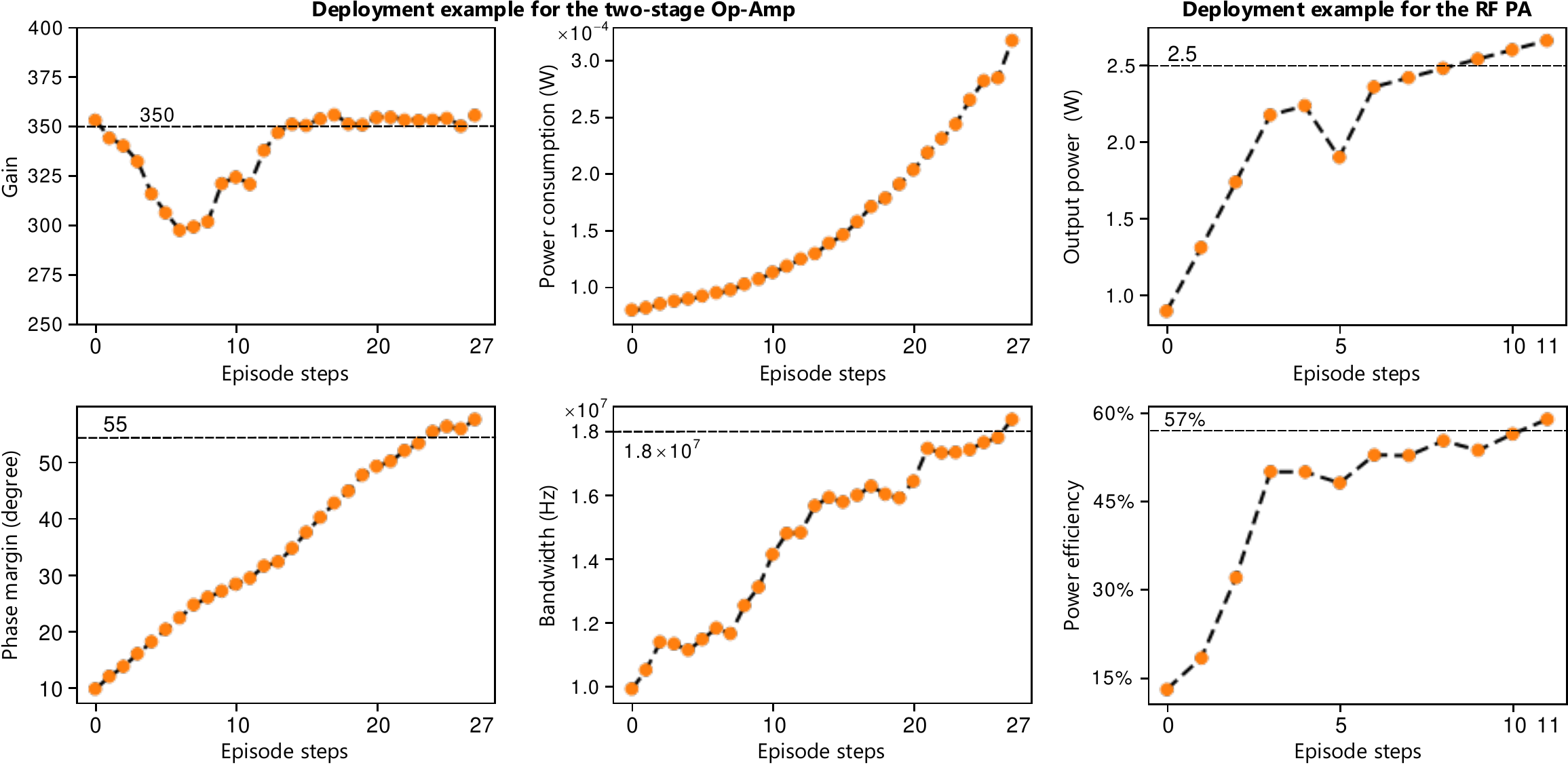}}
\caption{Deployment examples of the trained RL agent attempting to reach one group of the target specifications for each circuit.
The left column and the middle column correspond to the two-stage Op-Amp.
The right column corresponds to the RF PA.}
\label{fig: amp_traj}
\end{center}
\vskip -0.2in
\end{figure*}

\begin{figure*}[!t]
\begin{center}
\centerline{\includegraphics[width=0.75 \linewidth]{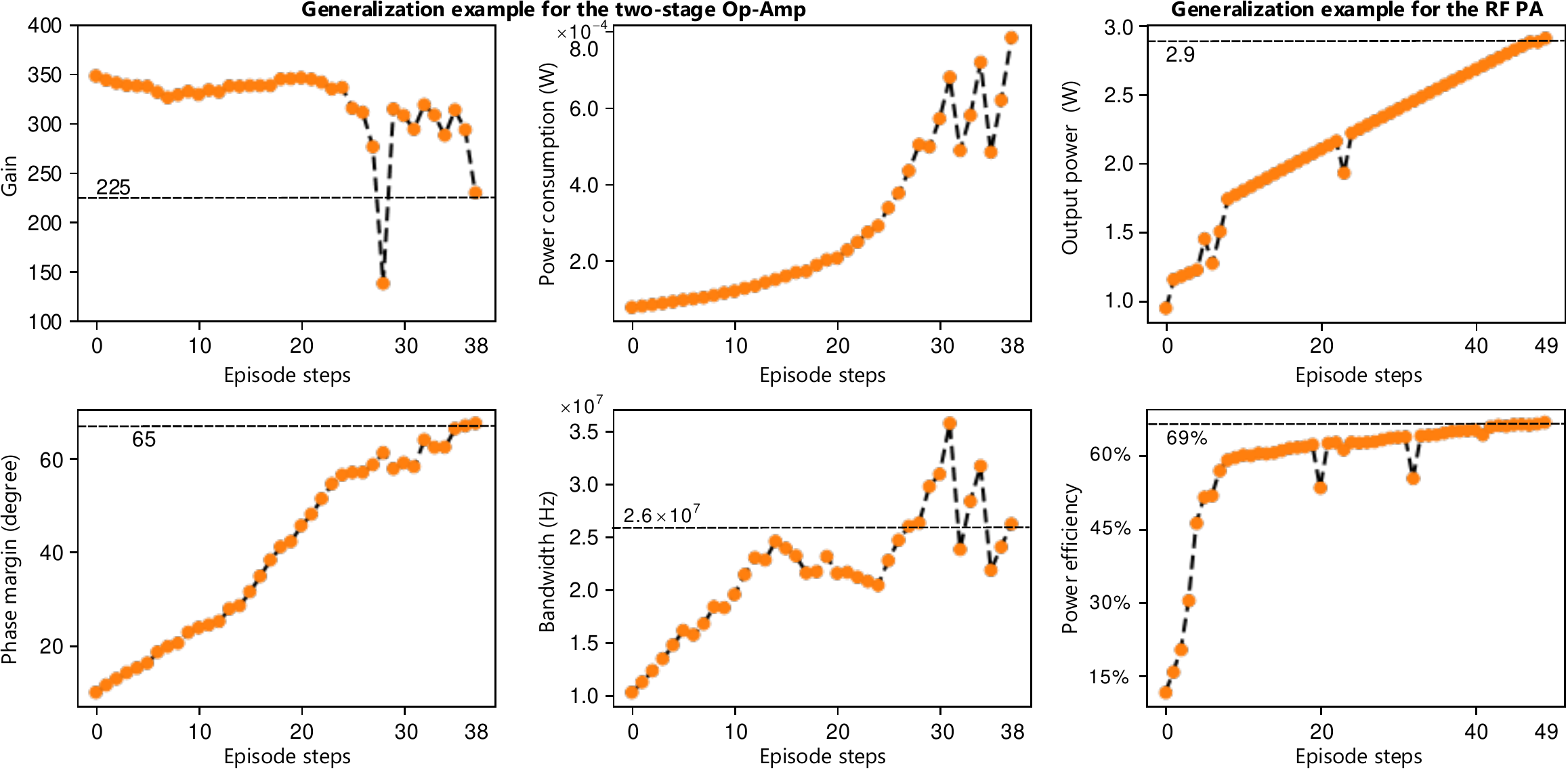}}
\caption{Generalization examples of the trained RL agent attempting to reach one group of the unseen new specifications for each circuit.
The left and middle column correspond to the two-stage Op-Amp.
The right column corresponds to the RF PA. }
\label{fig: gene}
\end{center}
\vskip -0.25in
\end{figure*}

\noindent{\textbf{Automated Design with Policy Deployment:}}
We take our GCN-FC-based policy as an example to show the deployment process.
Figure~\ref{fig: amp_traj} illustrates the deployment where RL agents automatically find optimal device parameters for a group of randomly sampled specifications (the horizontal dashed lines in each sub figure ).
The sampled desired specifications for the two-stage Op-Amp are gain ($G=350$), bandwidth ($B=1.8\cdot10^7$ Hz), phase margin ($PM= 55^{\circ}$), and power consumption ($P=4\cdot 10^{-3}$ W). 
And for the RF PA, they are output power ($P=2.5$ W), and power efficiency ($E=57\% $).
Note that the smaller the power consumption is, the better the performance is.
At the initial state, the intermediate specifications ($y$-axis of each sub-figure) often deviate a lot from the desired ones. 
As the deployment continues, they get closer to the desired ones by following the trained policy.
An interesting phenomenon here is that when some specification is first achieved, the RL agent will not over-optimize it too much but instead try to optimize the remaining ones.
For example, the gain of the two-stage Op-Amp is first attained at the $14^{\text{th}}$ deployment step. 
In the following steps, the RL agent focuses on optimizing phase margin and bandwidth.
The similar analysis also applies to the design of the RF PA.

We also analyze a few failed cases where our trained policy cannot converge to the optimal device parameters.
We observe that in these cases, some specifications can converge to a neighborhood of the desired ones, but after which they deviate a bit from the goal.
Fortunately, we find that by slightly tuning the device parameters with manual effort at that particular step, the design goal is also easily achieved.
In this way, the design accuracy can be improved to 100\%.
These results show that human designers can still greatly benefit from the trained policy, if used as an efficient warm-start of the manual tuning, even if an automated deployment fails.


\noindent{\textbf{Generalization to Unseen Specifications:}}
We also evaluate the generalization ability of our GCN-FC-based policy by deploying it with unseen specifications out of the sampling space in Table~\ref{tab1}.
Figure~\ref{fig: gene} shows such an example, where the horizontal dashed lines denote these unseen specifications: gain ($G=225$), bandwidth ($B=2.6\cdot10^7$ Hz), phase margin ($PM= 65^{\circ}$), power consumption ($P=6\cdot 10^{-3}$ W) for the two-stage Op-Amp; output power ($P=2.9$ W), and power efficiency ($E=69\%$) for the RF PA.
Compared to the policy deployment in Figure~\ref{fig: amp_traj} with the specifications coming from the sampling space, the deployment with unseen desired specifications usually requires more search steps.
For example, the generalization for the RF PA needs 49 steps to achieve the design goals while 11 steps are enough for the normal deployment in Figure~\ref{fig: amp_traj}. 
This is because that the unseen desired specifications are beyond the scope of the training datasets, thereby demanding more steps to reach the optimal parameters.
We also analyze the generalization ability of baseline methods and find that they often do not generalize well as ours even with a higher number of search steps.
The better generalization ability of our method is attributed to the fact that it is capable of capturing rich domain knowledge from state space, hence can leverage the learned experiences from previous states to the new unseen states at the inference time.


\begin{table*}[!t]
\caption{Comparison of different design automation methods.}
\begin{center}
\begin{scriptsize}
\vskip -0.15in
\begin{tabular}{c|c|c|c|c|c|c}
\toprule
\multirow{3}{*}{Methods} &
  \multicolumn{2}{c|}{\multirow{3}{*}{Sufficient key domain Knowledge (?)}} &
  \multicolumn{3}{c|}{P2S optimization} &
  FoM optimization \\ \cline{4-7} 
 &
  \multicolumn{2}{c|}{} &
  \multirow{2}{*}{\begin{tabular}[c]{@{}c@{}}Design\\ accuracy\end{tabular}} &
  \multicolumn{2}{c|}{Mean \# of design steps} &
  FoM value \\ \cline{5-7} 
 &
  \multicolumn{2}{c|}{} &
   &
  Two-stage Op-Amp &
  RF PA &
  RF PA \\ \midrule
Genetic Algorithm~\cite{genetic} &
  \multicolumn{2}{c|}{NO} & 76.7\%
   & 370
   & 389
   & 2.53
   \\ \hline
Bayesian Optimization~\cite{bayesian} &
  \multicolumn{2}{c|}{NO} & 83.7\%
   & 105
   & 86
   & 2.61
   \\ \hline
Supervised learning~\cite{op_amp_FCNN} &
  \multicolumn{2}{c|}{NO} & 79\%
   & 1
   & 1
   & N/A
   \\ \hline
\begin{tabular}[c]{@{}c@{}}RL method (Baseline A) \\~\cite{autockt_berke, RL_1}\end{tabular}&
  \multicolumn{2}{c|}{\begin{tabular}[c]{@{}c@{}}NO\end{tabular}} & 93\%
   & 27
   & 23
   & 2.92
   \\ \hline
\begin{tabular}[c]{@{}c@{}}RL method (Baseline B)\\ ~\cite{GCN_RL_MIT}\end{tabular} &
  \multicolumn{2}{c|}{\begin{tabular}[c]{@{}c@{}}NO. Implemented with our GCN (GAT) part\end{tabular}} & 84\% (87\%)
   & 32 (31)
   & 25 (23)
   & 2.81 (2.86)
   \\ \hline
\multirow{2}{*}{\textbf{Our RL method}} &
  \multirow{2}{*}{\begin{tabular}[c]{@{}c@{}}\textbf{YES. Full circuit topology}  + \\ \textbf{Specification couplings}\end{tabular}} & 
  \textbf{GCN + FCNN} & \textbf{98\%}
   & \textbf{24}
   & \textbf{19}
   & \textbf{3.18}
   \\ \cline{3-7} 
 &
   &
 \textbf{ GAT + FCNN} &\textbf{99\%}
   & \textbf{21}
   & \textbf{16}
   & \textbf{3.25}
   \\   \bottomrule
\end{tabular}
\end{scriptsize}
\label{tab2}
\end{center}
\vskip -0.10in
\end{table*}

\begin{figure}[!t]
\begin{center}
\centerline{\includegraphics[width=0.47 \linewidth]{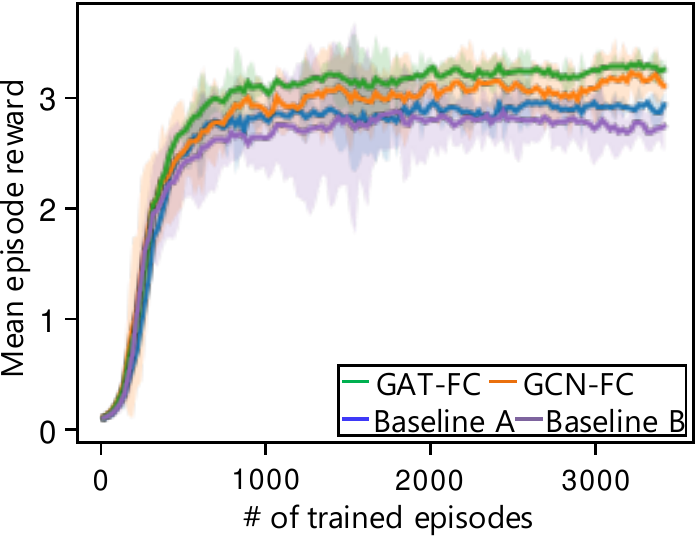}}
\caption{Comparing FoM optimization between different methods. All results are reported based on 6 random seeds.}
\label{fig: fom}
\end{center}
\vskip -0.30in
\end{figure}

\noindent{\textbf{FoM Optimization:}}
We also compare our method with the baselines~\cite{autockt_berke,RL_1,GCN_RL_MIT} in optimizing the FoM of analog circuits by using the RF PA as an example.
To apply all methods for this problem, we revise the reward function in Eq.~\eqref{eq:reward} based on the standard FoM definition of a RF PA in the prior work~\cite{bayesian}.
For each method, we train the corresponding RL agent with $3.5\times 10^3$ episodes.
Figure~\ref{fig: fom} shows the learning curves of different methods.
Our methods (GAT-FC/GCN-FC) can obtain a higher FoM.
And GAT-FC-based policy attains the highest one.
The comparisons still show the superiority of our methods in optimizing the FoM.

\noindent{\textbf{Comparison Summarization:}}
We summarize the comparisons between our method with all previous methods in terms of the two applications in Table~\ref{tab2}.
In tackling the P2S problem, our method achieves the highest design accuracy than all previous methods.
Optimization-based methods~\cite{genetic,bayesian} cannot ensure a high design accuracy because of their limitations, e.g., being stuck at a local optimum (caused by non convexity) or even divergence of the optimization algorithms. 
Due to the inherent approximation errors, supervised learning methods~\cite{op_amp_FCNN} suffer from a low design accuracy even including the domain knowledge.
RL methods~\cite{autockt_berke,RL_1,GCN_RL_MIT} without considering the main domain knowledge cannot reach the human-level design accuracy as ours.
Due to such limitations, these methods show a weaker generalization ability than ours, either.
Despite not excelling the design efficiency of supervised learning methods with one-step prediction, once trained our method uses fewer steps to find the optimal device parameters for the same desired specifications, improving the design efficiency by average $1.5\times$ as compared to the prior RL methods and average $10\times$ as compared to optimization-based methods.
In the application of FoM optimization, our method also achieves higher FoM value than prior RL methods and optimization-based methods.
In summary, our RL method inspired by key domain knowledge of analog circuit design and human-like multiple tuning steps achieves the best balance between design accuracy and efficiency and the best optimality.

\noindent{\textbf{Discussions:}}
Since GNN is a critical part of our method, we perform some discussions on several issues related to GNNs, e.g., cycles and scalability.
GNNs have limitations in detecting cycles in a graph.
Our method uses GNNs to capture the circuit physics not for the cycle detection, which may not suffer a lot from this limitation.
We also believe such a limitation would be resolved given the advancements in graph learning~\cite{GNN_limit}.
GNNs also have scalability issues when their sizes become large. 
The conventional philosophy to design a large-scale circuit is to decompose it into several small-scale essential building blocks, such as Op-Amps, filters, and mixers.
Our method applies to design large-scale analog circuits by providing a standard design methodology for such small-scale building blocks, which often have $\sim$10 or $\sim$100 order of devices (nodes).
Our method can be readily employed to design them with manageable scalability issue. 

\noindent{\textbf{Broad Impacts:}}
Our work could further impact both circuit design automation and deep learning.
Analog circuit design desires automation methods at all stages~\cite{Neural_PIM, Neu_ADC_TCAD, Pipeline_TCAD}.
Our work provides a solution to automate the design at the pre-layout stage.
However other stages, e.g., physical layout design, almost remain unexplored.
We think deep learning methods is promising to solve such problems.
Our method has shown the potential of GNNs in modeling circuits.
It could attract GNN community to develop insightful methods to better explain the connections between circuit topology and graph just like the analogy between molecular fingerprints and graph~\cite{graph_life}.
Last, combining domain knowledge of analog circuit design and deep learning may also inspire researchers in other fields to use similar principles to better solve domain-specific problems.
In a nutshell, we believe deep learning-based circuit design automation is an important rising field, which is worthy of strong explorations and interdisciplinary collaborations.



\section{Conclusion}
We have shown a deep RL method for automated design of analog circuits.
The key property of our framework is to incorporate domain knowledge of practical analog circuit design (i.e., the underlying physical topology of a given circuit and the trade-offs between specifications) into the newly proposed combined GNN (GCN/GAT)-FC-based policy network.
We show that such a method is superior to other methods without such considerations in designing various analog circuits with higher accuracy, efficiency, and optimality.
We expect that our method would assist IC industry to accelerate the analog chip design, with artificial agents that master massive circuitry optimization experiences via continuous training.

\noindent{\textbf{Acknowledgment:}} Weidong Cao was an intern at MERL. This work is supported by MERL with additional support for Weidong Cao, Xuan Zhang in part by National Science Foundation grant no. CCF-1942900.

\noindent{\textbf{Note:}} This paper is accepted by the 1st Annual AAAI Workshop on AI to Accelerate Science and Engineering (AI2ASE). A more advanced version is accepted by 2022 Design Automation Conference. The open source codes will be released soon. Stay tuned.

\bibliography{aaai22}

\end{document}